\documentclass[runningheads]{llncs}
\usepackage[T1]{fontenc}
\usepackage{graphicx}
\usepackage{xcolor}
\usepackage[hidelinks]{hyperref}
\usepackage{fontawesome5}
\usepackage{array}
\usepackage{multirow}
\usepackage{amsmath}
\usepackage{amsfonts}
\usepackage{makecell}
\usepackage{booktabs}
\usepackage{tabularx}
\usepackage{subcaption}
\usepackage{nth}
\usepackage{amssymb}

\definecolor{RankUp}{RGB}{34,139,34}
\definecolor{RankDown}{RGB}{178,34,34}

\newcommand{\triup}[1]{\textcolor{RankUp}{$\blacktriangle$\,#1}}
\newcommand{\tridown}[1]{\textcolor{RankDown}{$\blacktriangledown$\,#1}}
\newcommand{\trisame}{\textcolor{gray}{---}}

\newcommand{\qhdr}[1]{\textbf{#1}}

\usepackage{color}

\begin{document}

\title{Beyond Multilingual Averages: MTEB-PT, a Benchmark for Portuguese Sentence Encoders}

\titlerunning{MTEB-PT, a Benchmark for Portuguese Sentence Encoders}
%
\author{Lucas Hideki Takeuchi Okamura\orcidID{0000-0002-7198-6140} \and
Alexandre Alcoforado\orcidID{0000-0003-3184-1534} \and
Anna Helena Reali Costa\orcidID{0000-0001-7309-4528}}
\authorrunning{Okamura et al.}
%
\institute{Escola Politécnica, Universidade de São Paulo, São Paulo, Brazil \\
\email{\{lucasokamura, alexandre.alcoforado, anna.reali\}@usp.br}
}
\maketitle              
\begin{abstract}

Portuguese remains underrepresented in text embedding evaluation, despite being one of the most widely spoken languages in the world. As a result, embedding models are often selected based on English or multilingual metrics, while their effectiveness in Portuguese remains unclear. We present MTEB-PT, a Portuguese benchmark constructed from a subset of MMTEB, comprising 14 existing datasets across Semantic Textual Similarity (STS), classification, retrieval, and reranking. We use this benchmark to evaluate 17 open- and closed-source embedding models under a unified protocol. Our results show that Portuguese performance is strongly task-dependent: multilingual rankings do not reliably predict Portuguese-specific performance across task families, no single model dominates all settings, and models with stronger long-context capacity are particularly advantageous on longer-input tasks such as retrieval and reranking. The benchmark also shows that language-specific fine-tuning still improves model performance in Portuguese, especially on task types that match the adaptation data most closely. To examine this effect, we fine-tune three representative backbone models with Portuguese contrastive supervision and Matryoshka Representation Learning (MRL). These benchmark-informed baselines yield their strongest gains on STS, consistent with the predominantly symmetric supervision used during training, while also improving retrieval and remaining competitive under dimensional truncation. We release the MTEB-PT benchmark, the fine-tuned models, and the training and evaluation code.

\keywords{Text Embeddings \and Benchmark \and Portuguese \and Sentence Encoders \and Matryoshka Representation Learning}

\vspace{0.9em}










\end{abstract}

\section{Introduction}
\label{sec:introduction}

Text embeddings remain a core component of modern NLP, supporting tasks such as STS, classification, retrieval, and reranking. Even as decoder-based large language models have transformed generative NLP, encoder-based models are still often preferred when the goal is to obtain compact sentence representations with strong semantic structure for downstream use.

However, the evaluation of embedding models remains concentrated in English and in broad multilingual aggregates, making it difficult to assess progress for individual languages. This is particularly limiting for Portuguese: although it is one of the most widely spoken languages in the world, it remains comparatively underrepresented in standardized embedding evaluation. As a result, it is still unclear which models are genuinely strong for Portuguese, how rankings change across task families, and whether multilingual average performance translates into robust Portuguese performance.

To address this gap, we present MTEB-PT, a Portuguese benchmark slice of MMTEB~\cite{enevoldsen2025mmteb}, composed of 14 existing datasets spanning four task families: STS, classification, retrieval, and reranking. We curate a Portuguese evaluation setting within MMTEB and evaluate 17 open- and closed-source models under a unified protocol.

Our findings show that Portuguese embedding performance is strongly task-dependent. Models that perform well in multilingual settings do not transfer uniformly across Portuguese tasks, and the differences are especially visible between symmetric semantic tasks and longer-input retrieval or reranking settings, where long-context support is often advantageous. The benchmark also shows that language-specific fine-tuning remains useful in Portuguese. To examine this effect, we fine-tune three representative backbone models on Portuguese sentence-pair data and find the strongest gains on symmetric semantic tasks, especially STS, while still observing some transfer to retrieval.

Beyond full-dimensional evaluation, we also study dimensional efficiency in deployment-relevant settings. We train the adapted models with MRL~\cite{kusupati2022matryoshka}, which allows a single model to support progressive dimensional truncation and makes it possible to analyze how well Portuguese sentence representations retain quality under compact embeddings.

Our main contributions are:
\begin{itemize}
    \item \textbf{Portuguese benchmark:} We introduce MTEB-PT, a Portuguese benchmark constructed from a subset of MMTEB and composed of 14 existing datasets spanning STS, classification, retrieval, and reranking;
    \item \textbf{Benchmark findings:} We provide a systematic evaluation of 17 open- and closed-source embedding models and show that model rankings in Portuguese are strongly task-dependent, with multilingual strength not transferring uniformly across task families;
    \item \textbf{Strong Portuguese baselines:} We release fine-tuned versions of multilingual-e5-large, bertimbau-large, and modbertbr as benchmark-informed Portuguese baselines, showing that language-specific adaptation is especially beneficial for symmetric semantic tasks;
    \item \textbf{Dimensional efficiency analysis:} We show that MRL-trained models remain competitive under dimensional truncation, enabling more storage- and latency-efficient Portuguese embeddings.
\end{itemize}

\section{Related Work}
\label{sec:related_work}

Text embedding evaluation has increasingly moved from isolated semantic tasks to unified multi-task benchmarks. MTEB~\cite{muennighoff2023mteb} established a common evaluation framework across diverse embedding tasks, and MMTEB~\cite{enevoldsen2025mmteb} extended this setup to a broad multilingual setting. While these benchmarks greatly improve comparability across models, aggregate multilingual results still do not fully reveal language-specific behavior for languages such as Portuguese.

Prior work on Portuguese sentence embeddings has mostly evaluated narrower task subsets. Fialho et al.~\cite{fialho2020benchmarking} compared multilingual and Portuguese BERT-style models on NLI and STS using ASSIN, SICK-BR, and ASSIN2, while Gomes et al.~\cite{gomes2024openserafim} introduced Serafim encoders evaluated mainly on STS and retrieval. Peixoto et al.~\cite{peixoto2026global} show task-dependent behavior across Brazilian Portuguese classification, clustering, NLI, and STS, while Pinto~\cite{pinto2025matryoshka} studies Matryoshka training for dimensional efficiency. Concurrently, MTEB-BR~\cite{stekel2026mtebbrtextembeddingbenchmark} introduces a broader native Brazilian-Portuguese suite of 22 tasks across seven MTEB categories, evaluating 93 models and complementing MTEB-PT's MMTEB-derived slice with native retrieval and clustering coverage and uncertainty-aware leaderboard analysis.

Our work builds on these efforts by providing a single Portuguese benchmark slice within the MMTEB ecosystem, spanning STS, classification, retrieval, and reranking under a unified protocol.

\section{MTEB-PT: Portuguese Benchmark}
\label{sec:mteb-pt}

We introduce MTEB-PT, a Portuguese benchmark slice for evaluating sentence embedding models across multiple application settings. The benchmark follows the multi-task evaluation perspective of MTEB, which emphasizes that embedding quality is task-dependent and should not be inferred from semantic similarity performance alone~\cite{muennighoff2023mteb}. Rather than focusing only on STS, MTEB-PT includes four task families---STS, classification, retrieval, and reranking---so that models can be compared under both symmetric and asymmetric matching settings.

We selected datasets that satisfy three practical criteria: they are publicly available under open licenses, have sufficient quality for benchmarking, and are already accepted by the community through inclusion in the MTEB/MMTEB ecosystem. Since our goal is to provide a coherent Portuguese evaluation setting rather than introduce new datasets, we restrict the benchmark to Portuguese-labeled task subsets that can be evaluated under the official implementations.

\subsection{Construction and characteristics}
\label{sec:mtebpt:construction}

MTEB-PT contains \textbf{14 datasets} spanning four task families: \textbf{3 STS}, \textbf{5 classification}, \textbf{3 retrieval}, and \textbf{3 reranking}. The STS tasks evaluate pairwise semantic similarity; the classification tasks cover intent detection, sentiment analysis, toxicity, and hate speech; and the retrieval and reranking tasks extend the benchmark to asymmetric query--document settings relevant to semantic search and retrieval-augmented generation. Including retrieval and reranking is important because strong multilingual embeddings do not necessarily transfer uniformly from short symmetric tasks to longer and structurally asymmetric search scenarios.

Table~\ref{tab:mtebpt_main} summarizes the benchmark suite. In contrast to STS and classification tasks, retrieval and reranking datasets differ not only in supervision structure but also in evaluation scale, since they require ranking over full corpora or candidate sets. We therefore report their sizes in terms of query and document or candidate counts, which better reflects the actual evaluation setting. Most tasks involve relatively short texts, whereas the MultiLongDoc datasets define the benchmark's long-context regime and are particularly useful for stress-testing truncation sensitivity and long-document behavior.

\begin{table}[t]
\centering
\caption{Overview of the MTEB-PT benchmark.}
\label{tab:mtebpt_main}
\scriptsize
\setlength{\tabcolsep}{4.5pt}
\renewcommand{\arraystretch}{1.05}
\begin{tabular}{llll}
\toprule
\textbf{Task} & \textbf{Dataset} & \textbf{Subset} & \textbf{Evaluation size} \\
\midrule
\multirow{3}{*}{STS} & Assin2STS & -- & 2,448 pairs \\
 & SICK-BR-STS & -- & 2,048 pairs \\
 & STSBenchmarkMultilingual & pt & 1,379 pairs \\
\midrule
\multirow{5}{*}{Classification} & MassiveIntentClassification & pt & 2,974 examples \\
 & MultiHateClassification & por & 1,000 examples \\
 & BrazilianToxicTweetsClassification & -- & 2,048 examples \\
 & HateSpeechPortugueseClassification & -- & 2,048 examples \\
 & TweetSentimentClassification & portuguese & 870 examples \\
\midrule
\multirow{3}{*}{Retrieval} & WebFAQRetrieval & por & 10,000 q. / 209,353 docs \\
 & WikipediaRetrievalMultilingual & pt & 1,500 q. / 13,500 docs \\
 & MultiLongDocRetrieval & pt & 200 q. / 6,569 docs \\
\midrule
\multirow{3}{*}{Reranking} & WikipediaRerankingMultilingual & pt & 1,500 q. / 13,500 cand. \\
 & XGlueWPRReranking & pt & 677 q. / 8,314 cand. \\
 & MultiLongDocReranking & pt & 200 q. / 1,564 cand. \\
\bottomrule
\end{tabular}
\end{table}


\section{Experimental Setup}
\label{sec:exp_setup}

This section describes how we evaluate models on MTEB-PT under a unified and reproducible protocol. We present the evaluated models, the task-specific evaluation procedure, the prefix and formatting choices that ensure fair comparison, the aggregation strategy, and the dimensional truncation protocol. Throughout, our goal is to separate the definition of the benchmark from the choices made to run models on it consistently.

\subsection{Evaluated Models}
\label{sec:exp:models}

We compare four groups of models: \textbf{closed-source models}, \textbf{peer-reviewed open-source models}, \textbf{community open-source models} (not peer-reviewed), and \textbf{Portuguese-specialized models fine-tuned} in this work. This grouping reflects both practical usage and scientific lineage, allowing us to compare commercial APIs, widely studied research encoders, strong community baselines, and Portuguese-adapted variants under the same benchmark.

The fine-tuned group includes three backbone models --- {bertimbau-large}~\cite{souza2020bertimbau}, {multilingual-e5-large}~\cite{wang2024e5}, and {modbertbr}~\cite{wu2025modbertbr} --- which we adapt for Portuguese sentence-embedding learning using MRL. These backbones were selected to cover complementary starting points: a multilingual sentence encoder already strong in general embedding benchmarks, a Portuguese BERT-family model with larger capacity, and a lighter Portuguese encoder relevant for compact deployment settings. The training data, contrastive objective, and hyperparameter details are presented in Section~\ref{sec:pt_finetuning}.

Table~\ref{tab:model_metadata} summarizes all evaluated models, including model type, parameter scale, embedding dimensionality, and maximum context length. We report these properties explicitly because they are important for interpreting performance differences across task families, especially between short symmetric tasks and longer retrieval or reranking settings.

\begin{table}[tb]
\caption{Metadata of the evaluated models. SE = Sentence Encoder; PTE = Pre-Trained Encoder; MRL-SE = Matryoshka Representation Learning Sentence Encoder.}
\label{tab:model_metadata}
\centering
\scriptsize
\setlength{\tabcolsep}{4pt}
\renewcommand{\arraystretch}{1.05}
\begin{tabular}{lllrr}
\toprule
Model & Model Type & Model Size & \makecell{Embedding\\Dimension} & \makecell{Context Length\\(Tokens)} \\
\midrule

\multicolumn{5}{l}{\textbf{Closed-source}\vspace{3pt}} \\
cohere-embed-v4 & MRL-SE & -- & 3,072 & 128k \\
text-embedding-3-large & MRL-SE & -- & 1,536 & 8,192 \\
amazon-titan-v2 & MRL-SE & -- & 1,024 & 8,192 \\

\midrule
\multicolumn{5}{l}{\textbf{Peer-reviewed}\vspace{3pt}} \\
serafim-900m & SE & 0.9B & 1,536 & 128 \\
multilingual-e5-large & SE & 0.6B & 1,024 & 512 \\
multilingual-e5-large-instruct & SE & 0.6B & 1,024 & 512 \\
multilingual-e5-base & SE & 0.3B & 768 & 512 \\
multilingual-e5-small & SE & 0.1B & 384 & 512 \\
qwen3-embedding-0.6b & MRL-SE & 0.6B & 1,024 & 32k \\
gte-multilingual-base & MRL-SE & 0.3B & 768 & 8,192 \\
legal-bertimbau-large-sts & SE & 0.3B & 1,024 & 512 \\
multilingual-mpnet-base & SE & 0.3B & 768 & 128 \\
bertimbau-large & PTE & 0.3B & 1,024 & 512 \\
mmbert-base & PTE & 0.3B & 768 & 8,192 \\
modbertbr & PTE & 0.1B & 768 & 512 \\

\midrule
\multicolumn{5}{l}{\textbf{Community models}\vspace{3pt}} \\
mmbert-embed-32k & MRL-SE & 0.3B & 768 & 32k \\
neobertugues & PTE & 0.1B & 768 & 8,192 \\

\midrule
\multicolumn{5}{l}{\textbf{PT Fine-tuned MRL}\vspace{3pt}} \\
e5-large-matryoshka & MRL-SE & 0.6B & 1,024 & 512 \\
bertimbau-large-matryoshka & MRL-SE & 0.3B & 1,024 & 512 \\
modbertbr-matryoshka & MRL-SE & 0.1B & 768 & 512 \\

\bottomrule
\end{tabular}
\end{table}

\subsection{Evaluation Protocol}
\label{sec:exp:protocol}

We evaluate all models using the official MTEB/MMTEB task implementations and default evaluation procedures. This keeps MTEB-PT directly comparable to prior benchmark results while minimizing benchmark-specific modifications.

To ensure a fair comparison across heterogeneous architectures, we encode text using each model's recommended inference-time input format. For models explicitly trained with structured inputs—such as E5-family models that recommend the \texttt{query:} and \texttt{passage:} templates—we apply the corresponding prefixes during evaluation. For models without such documented conventions, we encode the raw text without adding prefixes. An exception is {e5-large-matryoshka} fine-tuned in this work: although the original model recommends using prefixes, our Portuguese fine-tuning was performed without them. Accordingly, we evaluate this fine-tuned variant without prefixes.

For \textbf{STS}, we compute cosine similarity between sentence embeddings and report Spearman's rank correlation with the gold similarity scores. For \textbf{classification}, we follow the standard MTEB protocol, training a logistic regression classifier on top of frozen embeddings using the task training split and evaluating on the corresponding test split. For \textbf{retrieval}, we rank the full document corpus for each query using cosine similarity and report nDCG@10, following the standard MTEB/BEIR retrieval setting~\cite{thakur2021beir}. For \textbf{reranking}, we score each candidate item with respect to the query embedding, sort the candidates accordingly, and report the metric defined by the official task implementation. In our benchmark, the metric is Mean Average Precision (MAP) for WikipediaRerankingMultilingual and XGlueWPRReranking, and nDCG@10 for MultiLongDocReranking, following the definitions established in the respective MMTEB task implementations. Although MAP and nDCG@10 differ in their formulations, both are bounded in $[0, 1]$ and capture ranking quality in comparable terms; macro-averaging across the three reranking datasets is therefore still meaningful as an aggregate indicator. Due to budget constraints, WebFaqRetrieval was run on only 20\% of the dataset for the closed-source models; results for this task should therefore be interpreted as partial.

Unless otherwise stated, we aggregate results by macro-averaging over datasets within each task family. We prefer task-family averages because they provide a more stable view of Portuguese model behavior than any individual dataset, while still preserving the distinction between symmetric semantic tasks and asymmetric search-oriented tasks.

\subsection{Dimensional Truncation Protocol}
\label{sec:exp:truncation}

Besides the main evaluation protocol, we perform a truncation analysis on the sentence encoders across multiple embedding sizes to examine how performance changes under compact representations and to assess the benefit of MRL for Portuguese.

This analysis is motivated by deployment constraints: full-dimensional performance is informative for leaderboard comparisons, but compact embeddings are often preferable in practical systems due to storage, memory, and latency considerations. By evaluating both regimes under the same benchmark, we can distinguish models that are strong only at maximum dimensionality from models that remain competitive after aggressive truncation.

For the truncation analysis, we extract lower-dimensional representations by selecting the first $d$ components of each model's full embedding. For MRL-trained models, this truncation is the intended usage mode; for non-MRL models, it serves as a baseline to determine whether structured multi-resolution training is necessary to retain quality at reduced dimensions.

\section{MRL Portuguese Fine-Tuning}
\label{sec:pt_finetuning}

Beyond benchmarking existing encoders, we study whether Portuguese-centered supervised adaptation still improves sentence embeddings under a modern multi-task evaluation setting. Our goal is not to propose a new architecture, but to build strong Portuguese baselines from representative backbones and test whether MRL preserves these gains under dimensional truncation.

We fine-tune three complementary backbone models: multilingual-e5-large, bertimbau-large, and modbertbr. The first is a strong multilingual sentence encoder that serves as a competitive cross-lingual baseline, whereas the latter two provide Portuguese-oriented masked-language-model backbones with different capacity and architectural trade-offs, including a lighter ModernBERT-based variant~\cite{warner2024modernbert}. For multilingual-e5-large, we use its native sentence-transformer pooling; for bertimbau-large and modbertbr, we add mean pooling over the final hidden states to obtain fixed-size sentence representations.

Our training data combines three complementary supervision sources: STS-style sentence pairs with graded similarity scores, Portuguese NLI entailment pairs, and retrieval-oriented positive pairs from MLDR~\cite{bge-m3}. Concretely, the STS portion includes Portuguese data from STSBenchmark~\cite{huggingface:dataset:stsb_multi_mt}, ASSIN~\cite{fonseca2016assin}, ASSIN~2~\cite{real2020assin2}, IRIS STS~\cite{legalbertimbau}, and SICK-BR~\cite{sickbr2018}, while the positive-pair portion uses Portuguese examples derived from MultiNLI, FeverNLI, ANLI, LingNLI, and WANLI~\cite{laurer_less_2022}. For all datasets, \textbf{only the official training splits are used during fine-tuning}; benchmark evaluation is conducted exclusively on the corresponding held-out test splits. This strict separation ensures that no evaluation pair is seen during training, preserving the integrity of benchmark comparisons. These models should therefore be interpreted as Portuguese-adapted reference baselines trained under a common recipe in this work, enabling a controlled view of how language-specific adaptation affects performance within MTEB-PT.

Training uses standard CoSENT~\cite{li2024cosent} for scored STS pairs and Multiple Negatives Ranking Loss~\cite{reimers2019sentencebert} for positive-pair supervision (NLI entailment and retrieval-oriented), both wrapped with MRL~\cite{kusupati2022matryoshka}. We apply the same training recipe across all three backbones for comparability and train each model so that it can be deployed at 64-, 128-, 256-, 512-, or full-dimensional embeddings. All models are trained for 20 epochs with an effective batch size of 512 (per-device batch size of 64 with gradient accumulation), a learning rate of \(5 \times 10^{-5}\), weight decay of~$0.2$, and a warmup ratio of~$0.1$.

\section{Results and Discussion}
\label{sec:results}

This section presents the main findings on MTEB-PT using full-dimensional embeddings, analyzes how model rankings differ from MMTEB when evaluation is restricted to Portuguese, and then examines how these patterns change under dimensional truncation.

\subsection{Main Benchmark Results}
\label{sec:results:main_results}

\begin{figure}[tb]
\includegraphics[width=\textwidth]{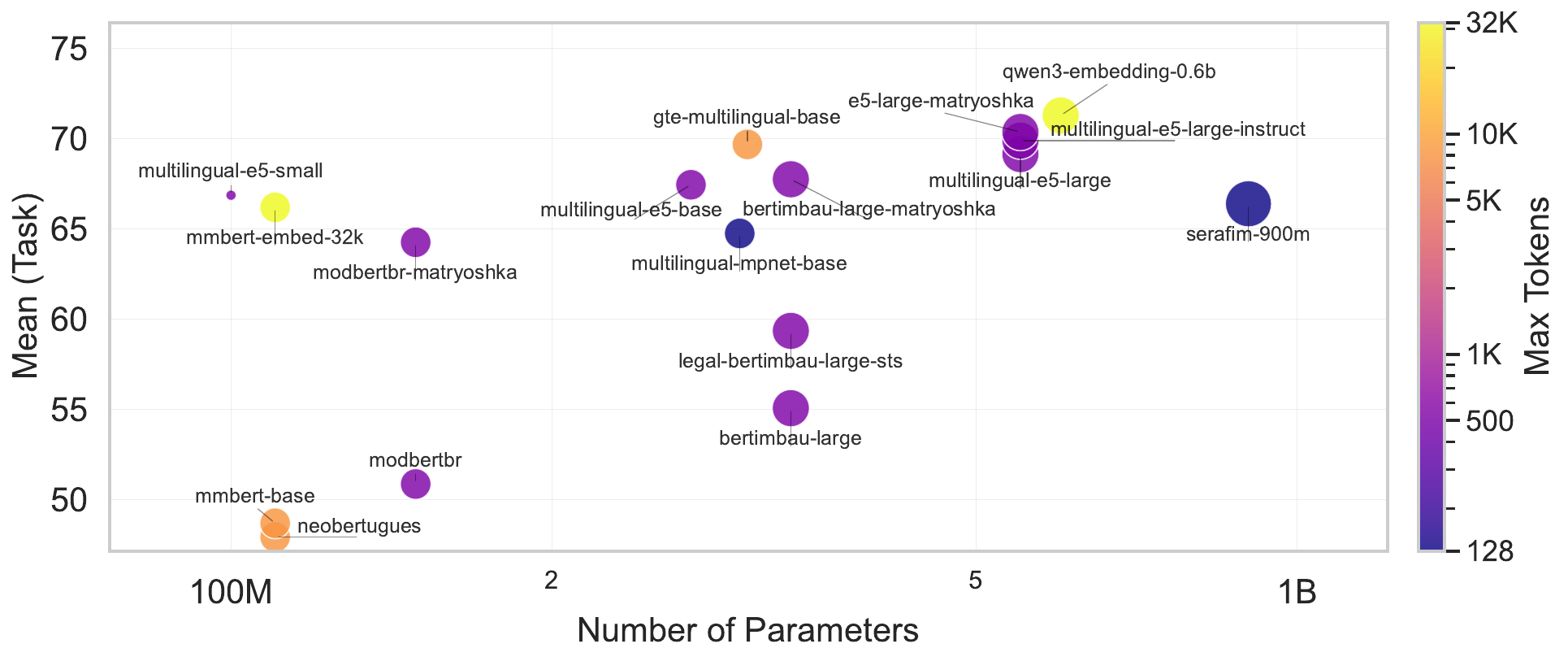}
\caption{Mean performance across tasks on MTEB-PT according to the number of parameters. The circle size denotes the full embedding size, while the color denotes the maximum sequence length of the model. Closed-source models are omitted because reliable parameter counts are not publicly available for all of them.}
\label{fig:global_results}
\end{figure}

\begin{table}[htb]
\caption{Results summary of the evaluated models across tasks, considering their full representation sizes (embeddings). *Closed-source retrieval results are approximate: \textit{WebFaqRetrieval} was evaluated on only 20\% of the dataset for these models, so their retrieval averages are not fully comparable to the other reported scores.}
\label{tab:model_results}
\centering
\scriptsize
\setlength{\tabcolsep}{4pt}
\renewcommand{\arraystretch}{1.05}
\begin{tabular}{lcccc}
\toprule
Model & STS & Classification & Retrieval & Reranking \\
\midrule

\multicolumn{5}{l}{\textbf{Closed-source}\vspace{3pt}} \\
cohere-embed-v4 & 77.9 & 51.4 & $\underline{90.6}^*$ & \textbf{86.9} \\
text-embedding-3-large & 79.1 & \textbf{56.3} & $\textbf{91.1}^*$ & 82.0 \\
amazon-titan-v2 & 77.0 & 50.2 & $86.9^*$ & 85.0 \\

\midrule
\multicolumn{5}{l}{\textbf{Peer-reviewed}\vspace{3pt}} \\
serafim-900m & \underline{86.7} & 52.3 & 54.9 & 81.0 \\
multilingual-e5-large & 79.2 & 51.3 & 73.4 & 83.8 \\
multilingual-e5-large-instruct & 81.3 & 53.0 & 70.1 & 84.5 \\
multilingual-e5-base & 76.9 & 49.1 & 71.7 & 84.5 \\
multilingual-e5-small & 76.9 & 49.3 & 70.2 & 84.0 \\
qwen3-embedding-0.6b & 80.7 & 52.3 & 78.9 & \underline{85.8} \\
gte-multilingual-base & 80.4 & 50.2 & 76.9 & 84.2 \\
legal-bertimbau-large-sts & 86.5 & 49.9 & 30.9 & 76.5 \\
multilingual-mpnet-base & 78.7 & 50.6 & 58.0 & 81.1 \\
bertimbau-large & 63.1 & 51.4 & 31.7 & 76.5 \\
mmbert-base & 59.4 & 48.0 & 17.6 & 70.3 \\
modbertbr & 51.2 & 49.4 & 27.2 & 76.7 \\

\midrule
\multicolumn{5}{l}{\textbf{Community Models}\vspace{3pt}} \\
mmbert-embed-32k & 76.4 & 49.5 & 67.3 & 82.7 \\
neobertugues & 57.0 & 48.8 & 18.0 & 67.5 \\

\midrule
\multicolumn{5}{l}{\textbf{Fine-tuned PT MRL}\vspace{3pt}} \\
e5-large-matryoshka & \textbf{88.0} & \underline{53.0} & 70.7 & 81.3 \\
bertimbau-large-matryoshka & 86.1 & 51.8 & 81.1 & 62.3 \\
modbertbr-matryoshka & 85.2 & 48.8 & 55.8 & 77.6 \\
\bottomrule
\end{tabular}

\end{table}

Figure~\ref{fig:global_results} summarizes how mean benchmark performance varies across open models as a function of parameter count, embedding size, and maximum context length, while Table~\ref{tab:model_results} shows where this aggregate view breaks down across task families. Closed-source systems are omitted from Figure~\ref{fig:global_results} because consistent model-size metadata is not publicly available.

Two broad patterns emerge from Figure~\ref{fig:global_results}. First, larger models tend to perform better on average, but the relationship is clearly not monotonic: models with similar parameter counts can differ substantially in mean performance, indicating that architecture and training objective matter at least as much as scale alone. Second, the strongest open models in the aggregate view are sentence-oriented encoders rather than raw pretrained language models, with {serafim-900m}, {qwen3-embedding-0.6b}, and {e5-large-matryoshka} occupying the upper portion of the plot despite representing different training paradigms and context-length regimes. This aggregate picture is informative, but it does not fully characterize Portuguese embedding quality. Table~\ref{tab:model_results} shows that performance remains strongly task-dependent, with substantial specialization across evaluation settings.

On \textbf{STS}, the strongest results come from Portuguese-adapted models. In particular, {e5-large-matryoshka} achieves the best overall STS performance, while {serafim-900m}, {legal-bertimbau-large-sts}, and {bertimbau-large-matryoshka} form a closely competitive group. This indicates that Portuguese-specific supervision still matters, especially for symmetric semantic tasks where sentence-pair ranking objectives are closely aligned with evaluation. Meanwhile, these same models do not consistently retain their advantage on search-oriented tasks, showing that strong STS performance should not be interpreted as evidence of uniformly strong Portuguese embeddings. This contrast is especially clear for \mbox{serafim-900m}, whose strong STS performance does not carry over to retrieval, likely reflecting the mismatch between short sentence-pair evaluation and longer retrieval documents under its 128-token context limit (Table~\ref{tab:model_metadata}).

The picture is different for \textbf{classification}. Here, the spread between models is smaller, and the leading systems are more general-purpose embedding models rather than Portuguese specialists. This suggests that probe-based classification is less sensitive than STS to fine-grained sentence-level alignment, and that reasonably strong semantic structure is often sufficient to obtain competitive performance. This also helps explain why Portuguese fine-tuning yields more modest gains on classification than on STS.

For \textbf{retrieval} and \textbf{reranking}, the benchmark reveals the strongest departure from the STS ranking. Closed-source models obtain the highest reported scores on these longer-input, asymmetric tasks, especially on retrieval, but the averages for these systems should be interpreted with caution because only 20\% of \textit{WebFaqRetrieval} was used for evaluating closed-source models (due to research budget and dataset size). Among open models, search-oriented multilingual encoders such as qwen3-embedding-0.6b, the stronger E5 variants, and gte-multilingual-base are substantially more competitive than STS-specialized Portuguese models. Taken together, these results suggest that multilingual strength does not transfer uniformly across task families, and that search-oriented training, often combined with longer context support, is advantageous on retrieval-style tasks.

The comparison between raw encoder backbones and sentence-trained models is also revealing. Portuguese MLM encoders such as {bertimbau-large} and {modbertbr} are weak in their off-the-shelf form, especially on STS and retrieval, despite being strong Portuguese language models in a pretraining sense. Once adapted with Portuguese contrastive supervision, however, their behavior changes substantially: {bertimbau-large-matryoshka} and {modbertbr-matryoshka} improve markedly on STS and also gain considerably on retrieval. This distinction is central to the benchmark: a strong Portuguese pretrained encoder is not necessarily a strong Portuguese sentence encoder, and sentence-level adaptation remains necessary for robust embedding quality.

The Portuguese MRL baselines therefore provide a more nuanced result than a simple ``fine-tuning works'' conclusion. Their largest gains are concentrated on STS, which is consistent with the predominantly symmetric supervision used during adaptation. Yet the adapted models also improve on retrieval, especially relative to their non fine-tuned backbones, indicating that Portuguese sentence-level supervision transfers beyond the task structure most directly targeted during training. In contrast, reranking remains dominated by models that appear more explicitly optimized for search-style matching.

Taken together, these results show that Portuguese embedding evaluation must remain task-diverse. Strong performance in the aggregate view of Figure~\ref{fig:global_results}, like strong performance on STS alone, is not sufficient to characterize model quality across Portuguese application settings.

\subsection{MMTEB vs. MTEB-PT}
\label{sec:mmteb_mteb-pt}

To further test whether multilingual benchmark performance transfers reliably to Portuguese, we compare model ranks on MMTEB and MTEB-PT across STS, classification, retrieval, and reranking. Because absolute scores are not directly comparable across benchmarks, we focus on within-task rank changes. Table~\ref{tab:quadrants-rank-mtebpt-vs-mmteb} reports model rankings on MTEB-PT and the corresponding rank differences relative to MMTEB. We include only models with available MMTEB results for all four task families.

\begin{table}[tb]
\caption{Models ranked by task on MTEB-PT. Each panel lists models sorted by their MTEB-PT rank. The rightmost value is the rank change relative to MMTEB: a green |$\blacktriangle$| indicates improved rank on MTEB-PT, a red |$\blacktriangledown$| indicates worse rank, and --- indicates no change. Only models with available MMTEB results for all four task families are included.}
\label{tab:quadrants-rank-mtebpt-vs-mmteb}
\centering
\scriptsize
\setlength{\tabcolsep}{6pt}
\renewcommand{\arraystretch}{1.1}

\begin{tabular}{p{0.47\linewidth} p{0.47\linewidth}}
\toprule

\qhdr{STS} & \qhdr{Classification} \\
\midrule

\begin{tabular}{@{}l l l@{}}
\nth{1} & multilingual-e5-large-instruct & \trisame \\
\nth{2} & qwen3-embedding-0.6b & \trisame \\
\nth{3} & gte-multilingual-base & \triup{1} \\
\nth{4} & multilingual-e5-large & \tridown{1} \\
\nth{5} & text-embedding-3-large & \triup{1} \\
\nth{6} & multilingual-mpnet-base & \triup{2} \\
\nth{7} & multilingual-e5-small & \tridown{2} \\
\nth{8} & multilingual-e5-base & \tridown{1} \\
\end{tabular}
&
\begin{tabular}{@{}l l l@{}}
\nth{1} & text-embedding-3-large & \triup{2} \\
\nth{2} & multilingual-e5-large-instruct & \trisame \\
\nth{3} & qwen3-embedding-0.6b & \tridown{2} \\
\nth{4} & multilingual-e5-large & \trisame \\
\nth{5} & multilingual-mpnet-base & \triup{3} \\
\nth{6} & gte-multilingual-base & \trisame \\
\nth{7} & multilingual-e5-small & \trisame \\
\nth{8} & multilingual-e5-base & \tridown{3} \\
\end{tabular}
\\

\midrule

\qhdr{Retrieval} & \qhdr{Reranking} \\
\midrule

\begin{tabular}{@{}l l l@{}}
\nth{1} & text-embedding-3-large & \triup{1} \\
\nth{2} & qwen3-embedding-0.6b & \tridown{1} \\
\nth{3} & gte-multilingual-base & \trisame \\
\nth{4} & multilingual-e5-large & \triup{1} \\
\nth{5} & multilingual-e5-base & \triup{1} \\
\nth{6} & multilingual-e5-small & \triup{1} \\
\nth{7} & multilingual-e5-large-instruct & \tridown{3} \\
\nth{8} & multilingual-mpnet-base & \trisame \\
\end{tabular}
&
\begin{tabular}{@{}l l l@{}}
\nth{1} & qwen3-embedding-0.6b & \triup{3} \\
\nth{2} & multilingual-e5-base & \triup{5} \\
\nth{3} & multilingual-e5-large-instruct & \trisame \\
\nth{4} & multilingual-e5-large & \tridown{2} \\
\nth{5} & gte-multilingual-base & \trisame \\
\nth{6} & multilingual-e5-small & \trisame \\
\nth{7} & text-embedding-3-large & \tridown{6} \\
\nth{8} & multilingual-mpnet-base & \trisame \\
\end{tabular}
\\

\bottomrule
\end{tabular}

\end{table}

The comparison shows that MMTEB rank is informative, but not sufficient, for Portuguese model selection. Some broad structure is preserved, since several strong multilingual models remain competitive on MTEB-PT. However, the relative ordering is not stable across task families, which indicates that multilingual performance does not transfer uniformly to Portuguese.

The amount of reordering is, itself, task-dependent. \textbf{STS} is the most stable setting, with only small rank changes and the top positions largely preserved. \textbf{Classification} shows moderate movement, suggesting that multilingual strength carries over only partially in this regime. Larger shifts appear in \textbf{retrieval}, where models with similar multilingual standing separate more clearly on Portuguese. The strongest mismatch appears in \textbf{reranking}, where several models change position substantially, including a marked drop for {text-embedding-3-large} and strong gains for {qwen3-embedding-0.6b} and {multilingual-e5-base}.

Overall, this comparison reinforces the main result of the benchmark: multilingual evaluation provides useful prior information, but it cannot replace Portuguese-specific evaluation. The extent to which multilingual performance transfers depends on the task family, with the largest mismatches appearing on the more search-oriented settings.

\subsection{Dimension Truncation Analysis}
\label{sec:results:dim_truncation}

\begin{figure}[htb]
  \centering
  \begin{subfigure}{0.5\linewidth}
    \centering
    \includegraphics[width=\linewidth]{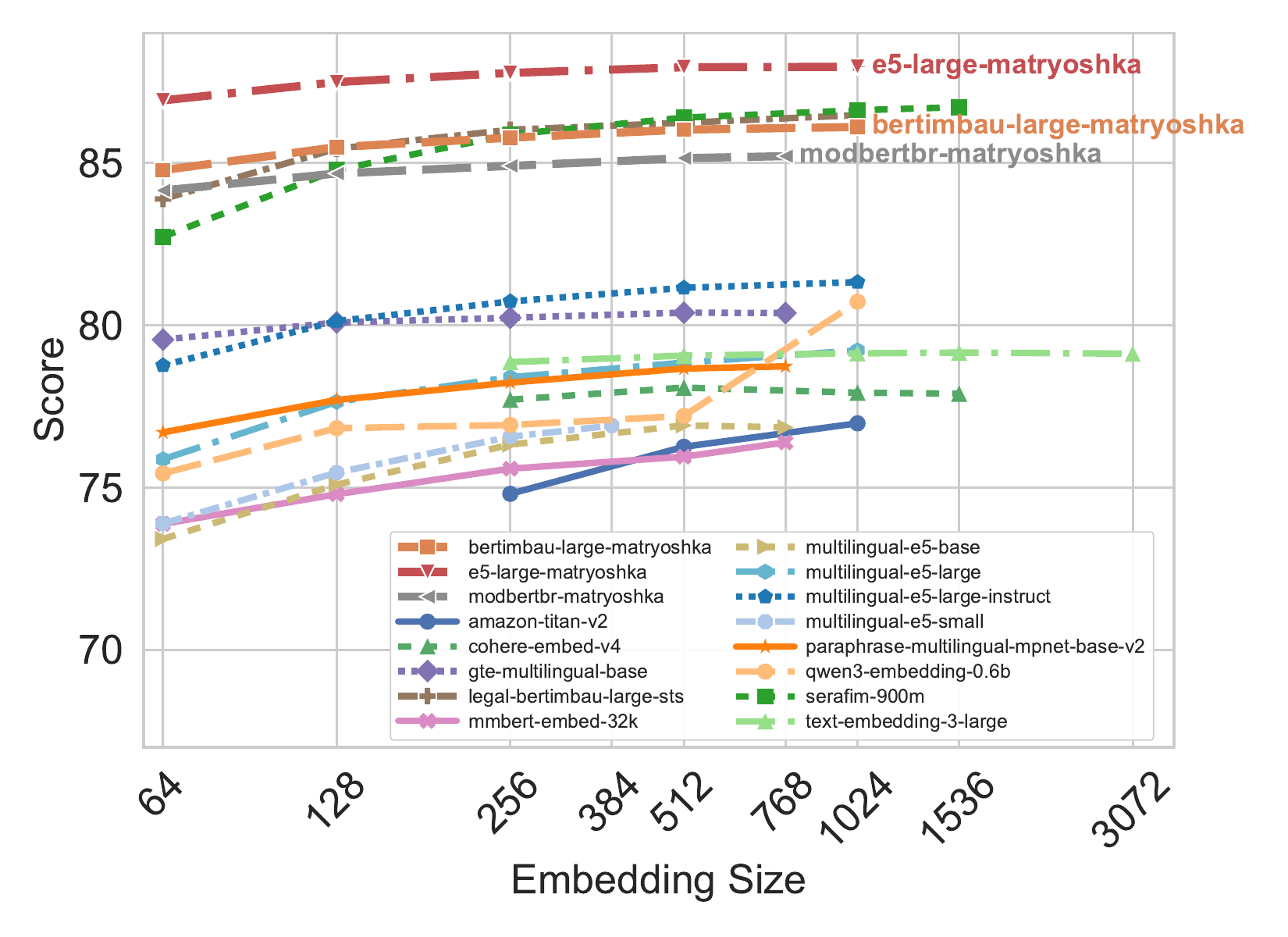}
    \caption{STS}
    \label{fig:tasks_dimensions:sts_task}
  \end{subfigure}\hfill
  \begin{subfigure}{0.5\linewidth}
    \centering
    \includegraphics[width=\linewidth]{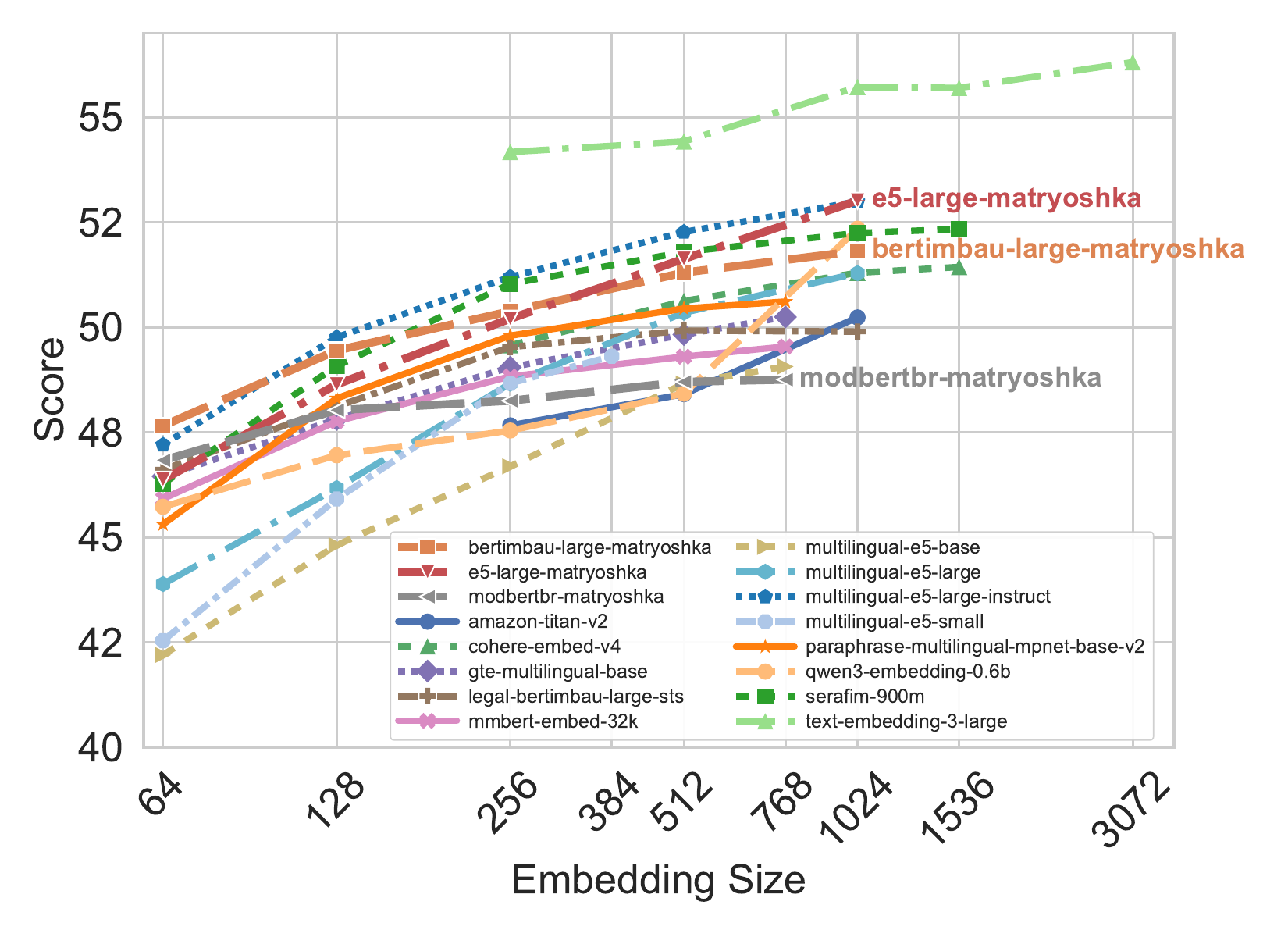}
    \caption{Classification}
    \label{fig:tasks_dimensions:classification_task}
  \end{subfigure}

  \begin{subfigure}{0.5\linewidth}
    \centering
    \includegraphics[width=\linewidth]{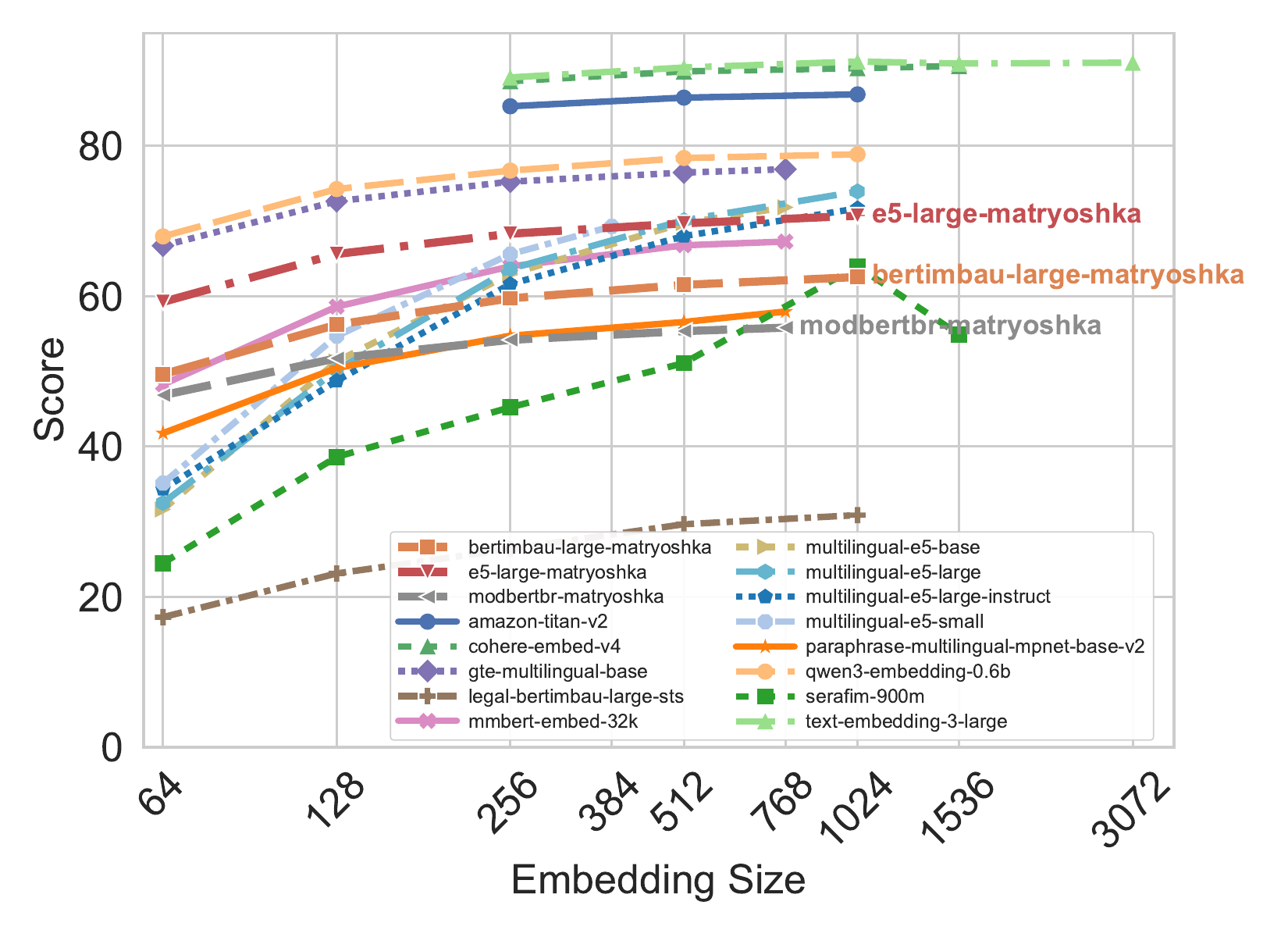}
    \caption{Retrieval}
    \label{fig:tasks_dimensions:retrieval_task}
  \end{subfigure}\hfill
  \begin{subfigure}{0.5\linewidth}
    \centering
    \includegraphics[width=\linewidth]{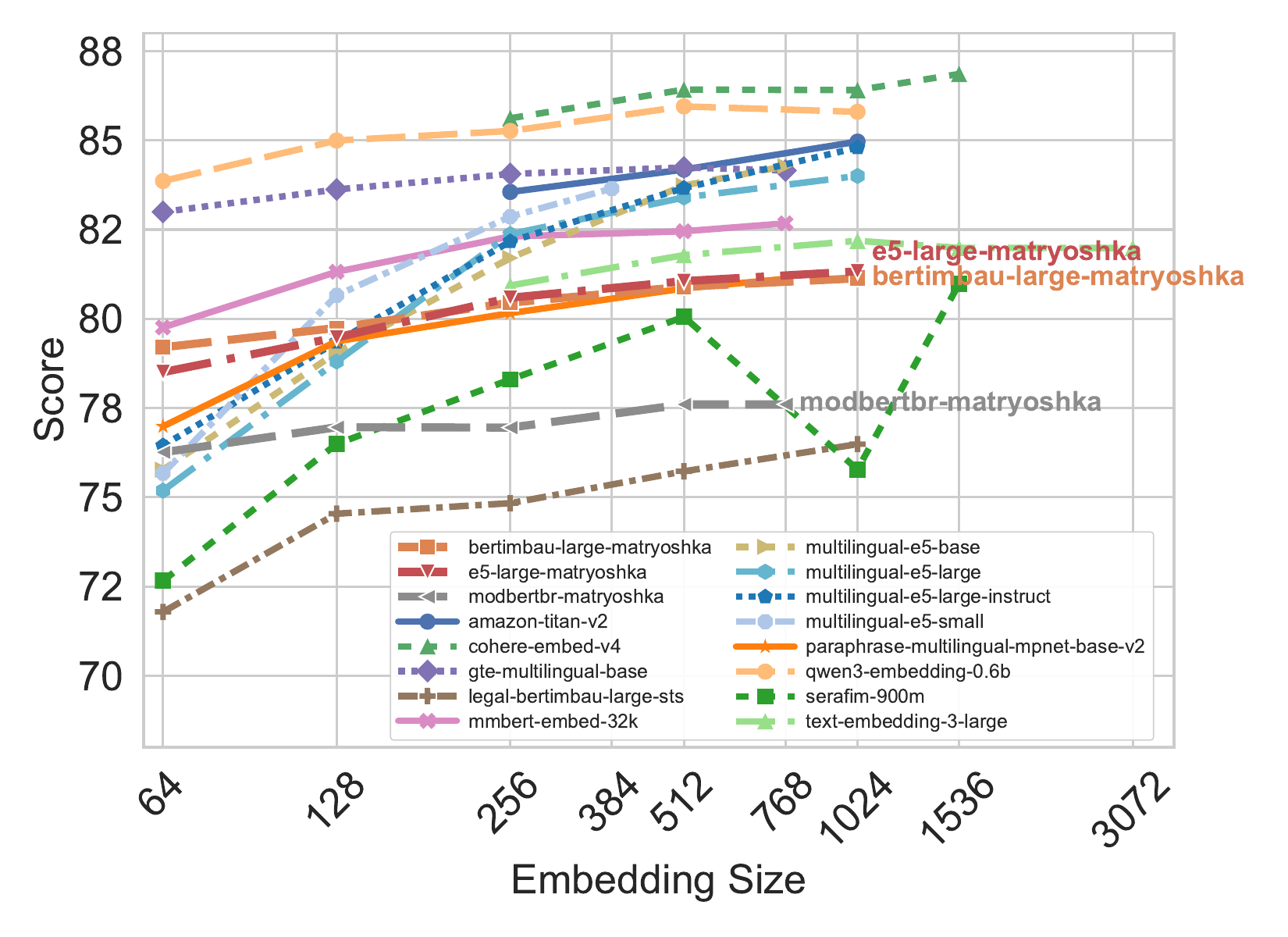}
    \caption{Reranking}
    \label{fig:tasks_dimensions:reranking_task}
  \end{subfigure}

  \caption{Task-wise performance under embedding truncation on MTEB-PT.}
  \label{fig:tasks_dimensions}
\end{figure}

We next examine how performance changes under dimensional truncation for sentence encoder models. Figure~\ref{fig:tasks_dimensions} shows that robustness to compression is also task-dependent: STS and classification remain comparatively stable, whereas retrieval and reranking are more sensitive to aggressive truncation.

The clearest MRL gains appear on \textbf{STS}, where the Portuguese baselines remain strong across dimensions, especially {e5-large-matryoshka}. \textbf{Classification} is flatter overall, with smaller gaps between models; here, the adapted models remain usable under compression, but their gains are more modest, consistent with the weaker alignment between pairwise semantic supervision and classification evaluation.

The hardest settings are \textbf{retrieval} and \textbf{reranking}. In both, closed-source systems and strong multilingual search-oriented encoders remain strongest overall, while the Portuguese MRL models improve substantially over their unfine-tuned backbones but do not displace the top search-oriented models. This is most evident for {e5-large-matryoshka} and {bertimbau-large-matryoshka}, which degrade more gracefully than their base encoders, indicating that Portuguese-specific adaptation transfers beyond STS even if it does not eliminate the advantage of models more directly optimized for search.

Overall, truncation does not remove the benchmark's main pattern: model quality remains task-dependent, and compact embeddings are easier to obtain for symmetric semantic tasks than for search-oriented ones. MRL mainly improves the practicality of Portuguese-adapted models under tighter dimensional budgets, with {e5-large-matryoshka} remaining the most balanced adapted model across task families.

\section{Limitations}
\label{sec:limitations}

MTEB-PT has several important limitations. First, it is a curated Portuguese slice of existing MTEB/MMTEB tasks rather than a fully comprehensive evaluation suite. Although it covers four task families, it does not yet capture the full diversity of Portuguese domains, genres, application settings, and linguistic phenomena. Notably, task families present in MTEB and MMTEB --- such as clustering and bitext mining --- are absent from MTEB-PT, not by design exclusion, but because no Portuguese-labeled subset meeting our quality and availability criteria was available within the MMTEB ecosystem at the time of construction. The absence of clustering is particularly relevant, as prior work on Brazilian Portuguese has shown that model behavior on grouping-based tasks can diverge substantially from STS and classification performance~\cite{peixoto2026global}.  Expanding the benchmark with additional task families, and especially with clustering datasets, is therefore an important direction for future work and would allow a broader and more representative evaluation of Portuguese sentence encoders.


Second, our evaluation focuses primarily on sentence encoders below one billion parameters, reflecting an emphasis on compact models that are feasible under modest memory and computational budgets. While this choice improves the practical relevance of the benchmark for low-resource deployment settings, it also limits the extent to which our results can speak to the performance of substantially larger embedding models. Evaluating larger models on MTEB-PT is therefore an important direction for future work, both to assess their actual gains on Portuguese and to determine whether those gains justify their additional computational cost relative to smaller encoders.

Finally, the interpretation of closed-source systems remains limited. These models are important practical baselines, especially on retrieval and reranking, but their training data and post-training procedures are not publicly documented. Also, reported averages are further limited by the fact that these models were evaluated on only 20\% of the \textit{WebFaqRetrieval} dataset, due to dataset size and research budget. Therefore, results on the retrieval task should be viewed as preliminary, although they already indicate data contamination (the dataset might have been used in the training of these models).

\section{Conclusion}
\label{sec:conclusion}

We introduced MTEB-PT, a Portuguese benchmark for sentence embeddings built from 14 datasets spanning STS, classification, retrieval, and reranking. By evaluating 17 open- and closed-source models under a unified protocol, we showed that Portuguese embedding performance is strongly task-dependent, that multilingual strength does not transfer uniformly across task families, and that models with stronger long-context or search-oriented capabilities are particularly advantageous on retrieval and reranking. Our comparison with MMTEB further shows that multilingual rankings provide useful prior information, but do not reliably predict Portuguese-specific model behavior across tasks.

We also used the benchmark to study Portuguese-specific adaptation through three MRL-trained baselines. These experiments show that language-specific fine-tuning remains effective, especially on STS, while also improving retrieval. Under dimensional truncation, the adapted models remain competitive, showing that compact Portuguese sentence representations can be obtained with favorable efficiency--performance trade-offs.

Overall, our findings reinforce the need for task-diverse, language-specific evaluation when selecting sentence encoders for Portuguese. We release MTEB-PT, the fine-tuned models, and the training and evaluation code to support reproducible research in Portuguese embeddings. We hope this benchmark encourages new datasets and broader Portuguese evaluation.

\begin{credits}
\subsubsection{\ackname} This study was financed in part by the Coordenação de Aperfeiçoamento de Pessoal de Nível Superior – Brasil (CAPES) – Finance Code 001. Anna H. Reali Costa would like to thank CNPq for the financial support, grant \#312360/2023-1, and the INCT TILD-IAR, grant \#408490/2024-1.
\end{credits}

%
%
%
%

\bibliographystyle{splncs04}
\bibliography{sample}






\end{document}